\documentclass[sigconf, screen]{acmart}

\usepackage{booktabs} 
\usepackage{subfigure}
\usepackage{enumitem}
\usepackage{amsmath}
\usepackage{hyperref}

\newcommand\blfootnote[1]{%
  \begingroup
  \renewcommand\thefootnote{}\footnote{#1}%
  \addtocounter{footnote}{-1}%
  \endgroup
}

\setcopyright{none}
\renewcommand\footnotetextcopyrightpermission[1]{}
\begin{document}
\title{3D Human Body Reshaping with Anthropometric Modeling}

\author{Yanhong Zeng}
\authornote{
  This work was performed when Yanhong Zeng was visiting Microsoft Research as a research intern.
}
\affiliation{
  \institution{School of Data and Computer Science, Sun Yat-sen University}
  \city{Guangzhou} 
  \postcode{510006}
  \country{P.R. China}
}
\email{zengyh7@mail2.sysu.edu.cn}

\author{Jianlong Fu}
\affiliation{
  \institution{Microsoft Research}
  \city{Beijing} 
  \postcode{100080}
  \country{P.R. China}
}
\email{jianf@microsoft.com}

\author{Hongyang Chao}
\affiliation{
  \institution{School of Data and Computer Science, Sun Yat-sen University}
  \city{Guangzhou}
  \postcode{510006} 
  \country{P. R. China}
}
\email{isschhy@mail.sysu.edu.cn}

\renewcommand{\shortauthors}{Zeng et al.}

\begin{abstract}
Reshaping accurate and realistic 3D human bodies from anthropometric parameters (e.g., height, chest size, etc.) poses a fundamental challenge for person identification, online shopping and virtual reality. Existing approaches for creating such 3D shapes often suffer from complex measurement by range cameras or high-end scanners, which either involve heavy expense cost or result in low quality. However, these high-quality equipments limit existing approaches in real applications, because the equipments are not easily accessible for common users. In this paper, we have designed a 3D human body reshaping system by proposing a novel feature-selection-based local mapping technique, which enables automatic anthropometric parameter modeling for each body facet. Note that the proposed approach can leverage limited anthropometric parameters (i.e., 3-5 measurements) as input, which avoids complex measurement, and thus better user-friendly experience can be achieved in real scenarios. Specifically, the proposed reshaping model consists of three steps. First, we calculate full-body anthropometric parameters from limited user inputs by imputation technique, and thus essential anthropometric parameters for 3D body reshaping can be obtained. Second, we select the most relevant anthropometric parameters for each facet by adopting relevance masks, which are learned offline by the proposed local mapping technique. Third, we generate the 3D body meshes by mapping matrices, which are learned by linear regression from the selected parameters to mesh based body representation. We conduct experiments by anthropomorphic evaluation and a user study from 68 volunteers. Experiments show the superior results of the proposed system in terms of mean reconstruction error against the state-of-the-art approaches. 
\blfootnote{This work was accepted as an oral paper in International Conference on Internet Multimedia Computing and Service 2017, Tsingtao, P.R. China (ICIMCS 2017).}
\blfootnote{
  The final publication is available at Springer via \url{https://doi.org/10.1007/978-981-10-8530-7_10}.
}
\end{abstract}

\begin{teaserfigure}
\center
  \includegraphics[width=0.9\textwidth]{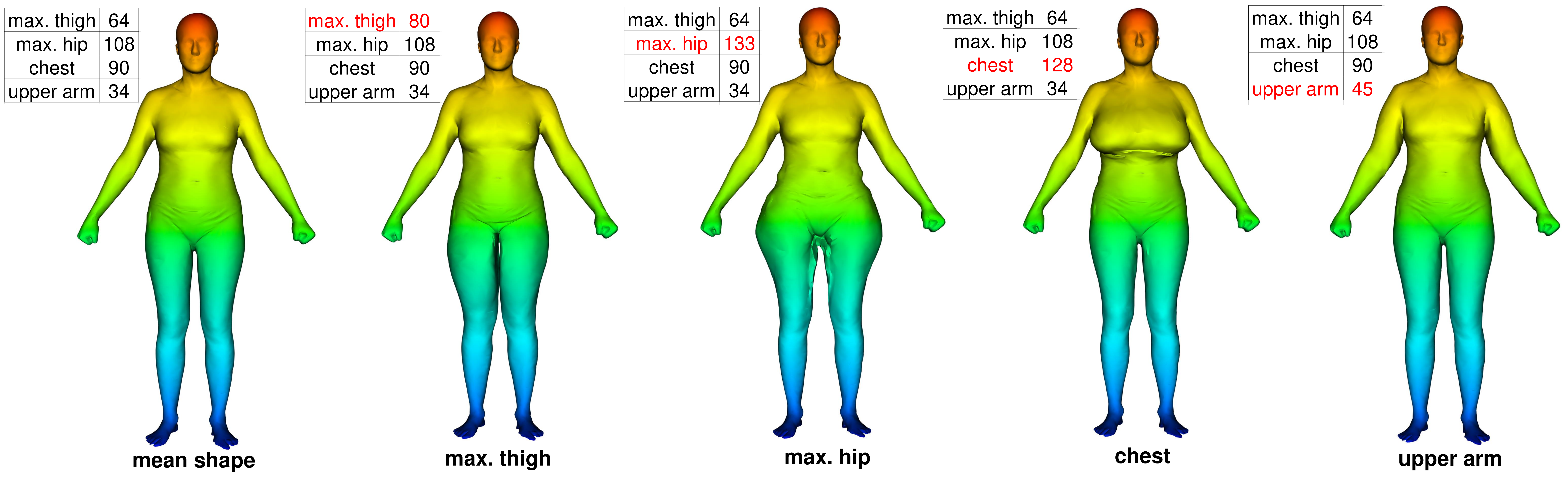}
  \caption{3D body reshaping results generated by the proposed system. Each body shape is generated by adjusting one of the anthropometric parameters (marked by red) to three times of standard deviation of a mean shape. For example, the fourth shape is generated by increasing the value of chest size while keeping other parameters unchanged. The four examplar parameters (cm) from 19 body measurements (in \autoref{tabel:parameters}) can be found in the top-left table for each shape. [Best viewed in color]}
  \label{fig:teaser}
\end{teaserfigure}

\maketitle

\section{Introduction}

Reshaping accurate and realistic 3D human bodies from anthropometric parameters plays a key role for real users, and can benefit a broad range of applications for person identification, health monitoring, online shopping and virtual reality. For example, reshaping an appropriate 3D human body can help people choose suitable clothes by fitting clothes into the 3D model, and thus the high refund rate for online shops and the waste of time for users can be largely reduced. However, building such a system is challenging, because the body shape of each person is unique, which is hard to be expressed by traditional anthropometric parameters (e.g., height, weight, chest size, etc.) even for the users themselves.

Significant progress has been made by introducing datasets on 3D human body reshaping (e.g., CAESAR dataset \cite{robinette_civilian_2002}), which provides opportunities for representing human bodies by 3D models and motivates great number of research works on learning statistical models for 3D human bodies. These works mainly focus on learning linear regression models from anthropometric parameters and PCA coefficients from vertex information in 3D body shapes \cite{allen_space_2003, seo_automatic_2003, allen_exploring_2004}. Furthermore, Sumner et al. proposed to reformulate the deformation transfer problem in terms of the triangles' vertex positions in their work \cite{sumner_deformation_2004}. Such an approach introduced more complex models for 3D body reshaping by a deformation transfer method for the first time. Later, Y. Yang et al. proposed the SPRING model, which outperforms deformation-based global mapping methods with a significant margin \cite{yang_semantic_2014}.

However, the SPRING model requires segmenting the body shape into fixed rigid parts beforehand (e.g., 16 parts) and binding anthropometric parameters to each part manually for calculating mapping relationship, which brings several limitations. First, defining the relationships among these parts and more than 20 parameters manually often involves heavy human efforts. Second, the definition can vary a lot among different people, which may bring inconsistent results using the same algorithm. Third, binding parameters to facets by hand totally loses the statistic information among the parameters, which may cause some error-prone results. Moreover, creating such 3D shapes from range cameras \cite{bogo_detailed_2015, li_3d_2013} or high-end scanners \cite{allen_learning_2006,anguelov_scape:_2005,loper_smpl:_2015} often requires complex measurement devices that are not widely available. A high precision 3D scanner usually costs thousands of dollars, which limits the applications for common users in real scenarios.

\begin{figure}
\includegraphics[width=\columnwidth]{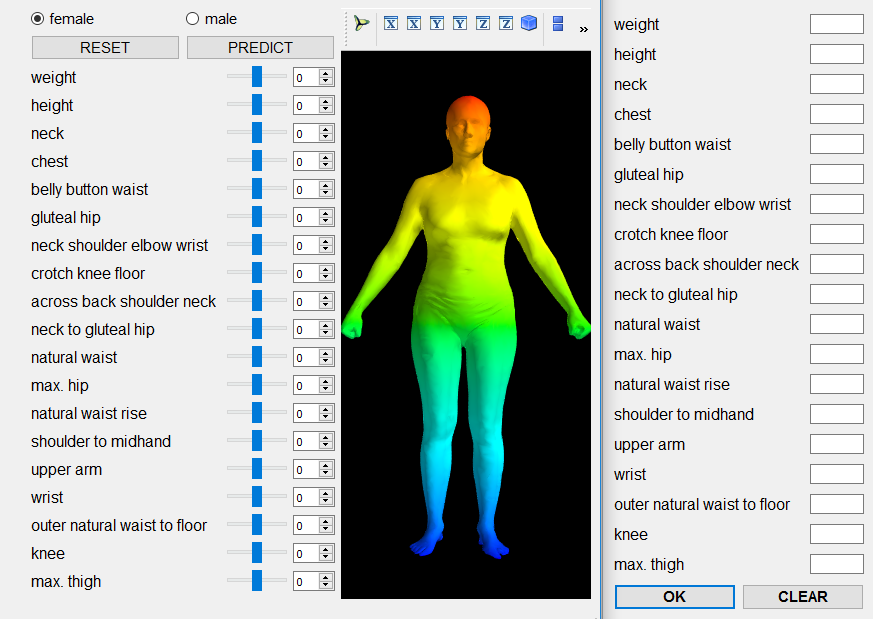}
\caption{A screenshot of the proposed system. User can (1) enter a few anthropometric parameters in the right panel; 
(2) view the generated body shape inside the center window; 
(3) refine the current body shape slightly by using the slide controls on the left panel as they expected.} 
\label{Fig:demo}
\end{figure}

To address the above problems, we have designed a 3D human body reshaping system by proposing a novel feature-selection-based local mapping technique with limited user inputs. First, we propose to adopt Multivariate Imputation by Chained Equations (MICE) technique to fill in the missing parameters from the limited user's input \cite{ azur_multiple_2011}, which improves the precision of model and enables better user experience in real scenarios. Second, the proposed feature-selection-based local mapping method takes advantage of recursive feature elimination techniques to remove irrelevant anthropometric parameters recursively based on the linear regression weights from a parameter set to mesh based body representation for each facet \cite{guyon2002gene}. As a result, an optimal relevant subset of features will be selected automatically, which eliminates the heavy human involvement from manually binding parameters into body parts. Third, we conduct anthropomorphic evaluation and a user study from 68 volunteers for the proposed system. From these experiments, we find that the proposed system outperforms the state-of-the-art approaches with a clear margin. \autoref{Fig:demo} shows an examplar screenshot of our system. The contributions of this paper are summarized as follows:
\begin{itemize}[nosep]
\item Our system provides users with an accurate body shape by using limited parameters as inputs, which ensures better user experience in real applications. 
\item We propose a novel feature-selection-based local mapping method for anthropometric modeling, which can jointly bind an optimal relevant subset of parameters to each facet.
\item We design comprehensive experiments from objective and subjective aspects to show the superiority of our system.
\end{itemize}

The rest of the paper is organized as follows. Section 2 introduces the proposed system. Section 3 describes the evaluation and analysis, followed by the conclusion in Section 4.

\section{System}

\begin{figure*}
\includegraphics[width=\textwidth]{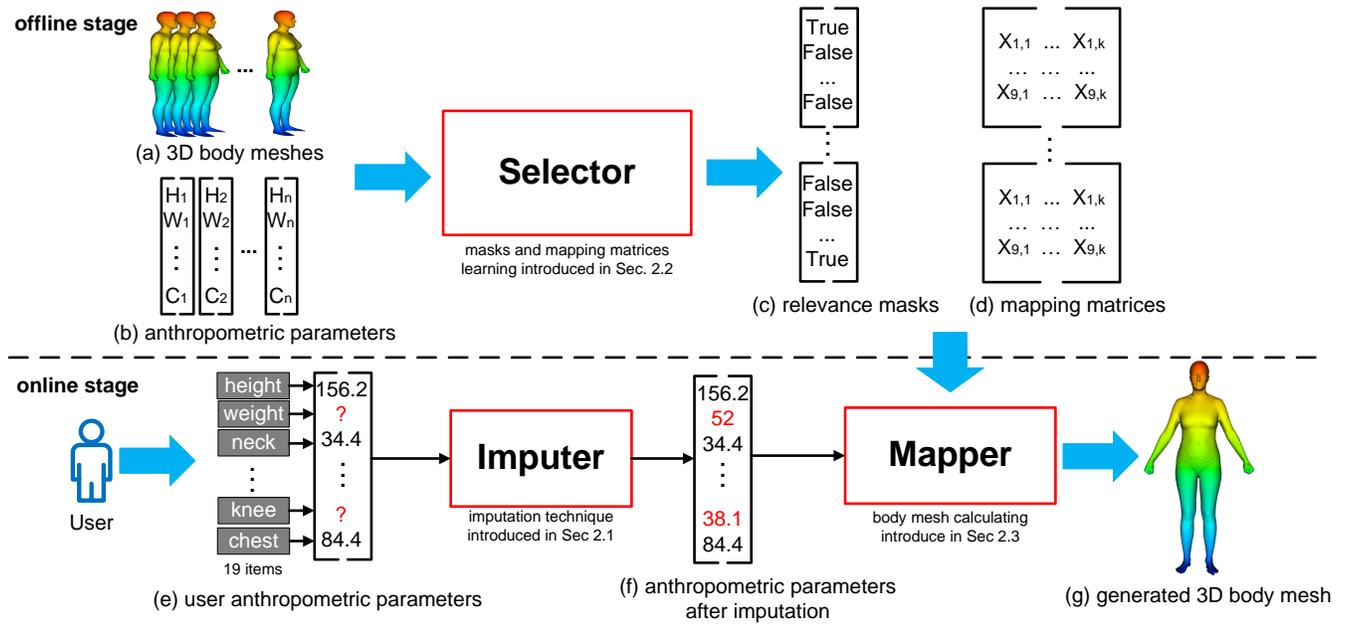}
\caption{The overview of the proposed 3D human body reshaping system. The system consists of three parts, i.e., the Imputer, the Selector and the Mapper in both online and offline stage. In offline stage, the Selector takes the dataset of 3D body meshes (a) and corresponding anthropometric parameters (b) as inputs to learn the relevance masks (c) by the proposed feature-selection-based local mapping technique. The mapping matrices (d) are further learned from the parameters selected by (c) to mesh based body representation. The details of the Selector are introduced in Sec. 2.2. In online stage, MICE is leveraged in the Imputer for the imputation of user inputs (e), which is introduced in Sec. 2.1. "?" in (e) indicates the missing parameters from user inputs, yet could be complemented in (f) by MICE. After imputation, the vector of parameters (f) will be passed to the Mapper. By adopting (c) and (d), 3D body mesh (g) will be generated from (f) in the Mapper, which is introduced in Sec. 2.3.} 
\label{Fig:System}
\end{figure*}

To develop an accurate and user-friendly 3D human body reshaping system, we propose to leverage MICE and propose the feature-selection-based local mapping method for anthropometric modeling in our system. Specifically, our system consists of three modules in offline stage and online stage, i.e., the Imputer, the Selector and the Mapper. The overview of our system can be found in \autoref{Fig:System}. Several relevant data are marked as (a-g) in \autoref{Fig:System}. 

The Imputer module is primarily responsible for imputation of user's input ( (e) in \autoref{Fig:System}), which may contain missing data. After imputation, the vector of parameters (f) will be passed to the Mapper module. The Mapper module selects the most relevance parameters from (f) by adopting relevance masks (c) for each facet, and then generates a final body mesh (g) by mapping matrices (d). (c) and (d) are learned offline by the proposed feature-selection-based local mapping technique in the Selector module. The Selector module takes the dataset of body mesh (a) and corresponding anthropometric parameters (b) as inputs to learn (c) and (d) offline. The details of these modules will be introduced in 3.1, 3.2 and 3.3.


\subsection{Imputer}

\begin{table}
\caption{19 parameters used in our models. These parameters have covered primary body measurements, and each item has a strict anthropometric definition (e.g. natural waist means the minimum circumference of the mid torso). More details can be referred in the Measurement Guide \cite{noauthor_bodylabs_nodate}.} 
\label{tabel:parameters}
\begin{tabular}{|c|c|c|c|} \hline
1 &weight &2 &height \\ \hline
3 &neck &4 &chest \\ \hline
5 &belly button waist &6 &gluteal hip \\ \hline
7 &neck shoulder elbow wrist &8 &crotch knee floor \\ \hline
9 &across back shoulder neck &10 &neck to gluteal hip \\ \hline
11 &natural waist &12 &maximum hip \\ \hline
13 &natural waist rise &14 &shoulder to midhand \\ \hline
15 &upper arm &16 &wrist \\ \hline
17 &outer natural waist to floor &18 &knee \\ \hline
19 &maximum thigh & \ &\ \\ \hline
\end{tabular}
\end{table}

Since more anthropometric parameters contained in the reshaping model, more constrains will be used when generating the 3D body mesh, which results in a more approximate body mesh. Our system allows users to enter as up to 19 items of anthropometric parameters as listed in \autoref{tabel:parameters} (e.g., height, weight, chest etc.) so that more approximate body meshes can be obtained. In case most users do not remember all values of these 19 parameters and the input may contain missing data, the Imputer module needs to preprocess the parameters so that a completed 19 dimensional vector of parameters can be passed to the Mapper module for further calculation.

Imputation is a task to predict and interpolate the missing item in dataset with statistical analysis of dataset. The most common imputation techniques are mean substitution method, similarity coefficients simple average method, K nearest method, MICE etc. \cite{ su_survey_2009}. Specifically, MICE is considered to be one of the best ways to impute missing data. To simulate the uncertainty of missing data, Rubin et al. established this method by generating a series of possible values instead of each missing value \cite{ royston_multiple_2004}, and the results of each missing data will be predicted after a standard statistical analysis on the generated datasets.

In order to choose the optimal imputation technique for the Imputer module, we have designed experiments to evaluate the above four imputation techniques. We have 19 parameters in our system, which means that there are $2^{19}$ possible input cases. To try out all these possible situations is meaningless. So we designed five sets of experiments, which involve most common situations and most important parameters used in \cite{ seo_example-based_2004}. Under the same time consumption, we found that MICE performs best in our experiments, so we choose MICE as the core technique of the Imputer. In our system, the Imputer takes the parameters from user input together with our datasets to run MICE process for imputation.


\subsection{Selector}

The Selector module takes the datasets of 3D body meshes and corresponding anthropometric parameters as inputs to learn the relevance masks and mapping matrices for each facet by the proposed feature-selection-based local mapping method.

The analysis of 3D human body shape can be divided into two paradigms as point based and mesh based \cite{tsoli2014modeling}. Our system adopts the latter one, which analyzes the deformations of each triangle facet. Mesh based analysis can factor out other variations (e.g. pose etc.) in some complex models and enables us to use local mapping method in simple body reshaping models. The deformations of each triangle facets can be computed using the similar method in \cite{sumner_deformation_2004}.

After the deformations across of all bodies in the dataset are obtained, we denote the deformation of each facet in each body mesh as a $3 \times 3$ transformation matrix:
\begin{equation}
Q = 
\begin{bmatrix}
q_{1,1} &q_{1,2} & q_{1,3} \\ 
q_{2,1} & q_{2,2}  & q_{2,3} \\ 
q_{3,1} & q_{3,2}  & q_{3,3} 
\end{bmatrix},
\end{equation}
then the deformations of body mesh $i\in 1, \cdots, n$ can be expressed as $S_i = \left[ Q_{i,1}, Q_{i,2}, ..., Q_{i,m} \right] ^T $, where $m$ is the number of facets in a body mesh, $n$ is the size of datasets. For each body mesh in our datasets, there is a 19 dimensional vector of anthropometric parameters value $P$, which are extracted from the body mesh by point-to-point distance with sets of control points. 

SPRING proposed to segment the body mesh into 16 rigid parts, then bind a vector of the most relevant parameters ${P}'$ from $P$ to the parts. It learns the linear regression model between deformation matrix $Q$ and the relevant parameters ${P}'$ for each facets directly, which utilizes the topology information of the human body shape and outperforms global mapping method with a large margin \cite{ yang_semantic_2014}. Such an approach has limits in binding parameters to the parts manually, which results in inconsistent results by different binding set and involves heavy human efforts .

To address the problems above, we propose the feature-selection-based local mapping method, which can bind parameters to the parts automatically. There is no need for our method to segment the body mesh and specify the relevant parameters for each facet. Our method propose to analyze the statistic information between parameters and deformations to select the most relevant parameters for each facets automatically. The details are introduced as follows.

Consider only one facet. Here we denote the matrix of anthropometric parameters for n body meshes as
\begin{equation}
X = 
\begin{bmatrix}
p_{1,1} &\cdots & p_{1,19} \\ 
\vdots & \cdots  & \vdots \\ 
p_{n,1} & \cdots  & p_{n,19} 
\end{bmatrix},
\end{equation}
where $p_{i,j}$ means the $j$th ($j\in 1, \cdots, 19$) parameter of body $i \in 1, \cdots, n$. Since $Q$ is a linear transformation matrix, the absolute value of the determinant of $Q$, $det(Q)$, can reflect how much the transformation expands the "volume" of the facet, so we can calculate the determinant of each transformation matrices as a feature of each deformation. The determinants of transformation matrices of a facet for n body meshes is given by:
\begin{equation}
Y = 
\begin{bmatrix}
det(Q_1)  \\ 
det(Q_2) \\ 
\vdots  \\
det(Q_n)
\end{bmatrix}
=
\begin{bmatrix}
d_1  \\ 
d_2 \\ 
\vdots  \\
d_n
\end{bmatrix}.
\end{equation}

For a facet, our method learns linear regression between $X$ and $Y$ to select the most important parameters from 19 items by recursive feature elimination algorithm \cite{guyon2002gene}. The recursive feature elimination algorithm learns linear regression on the whole set of features at first. Then, features with smallest absolute weights will be pruned and a new linear regression model is then retrained until the desired number of features to select is eventually reached. Our method binds the optimal relevant subsets of parameters in this way, at the same time a vector of corresponding relevance mask will be generated. If the parameter is selected for this facet, then the label in the vector of this facet will be true, otherwise it will be false. After selection, we reshape the top $k$ relevant parameters as a $k$ dimensional vector ${P}'$. Then a linear regression model can be learned from ${P}'$ to a transformation matrix $Q$. The mapping matrix $M$ for a facet is a $9 \times k$ dimensional. From experiments, we found that, the results of feature selection are consistent with the topology knowledge of human body shape.


\subsection{Mapper}

The Mapper module synthesizes a final 3D human body mesh for users in our system. It maps the 19 dimensional vector of anthropometric parameters after imputation to a mesh based body representation, which consists of a list of deformations for each facets. The vertex positions of the mesh based representation can be computed from the deformations using the method in \cite{sumner_deformation_2004}. The details of this module are introduced as follows.

Specifically, the Mapper module consists of two steps. First, the Mapper module selects a subset of parameters for each facet from the 19 dimensional vector of anthropometric parameters by adopting the relevance masks, which are learned in the Selector (details in Sec 2.2). It reshapes the selected $k$ parameters ($k < 19$) as a k dimensional vector ${P}'$. Second, we obtain the deformation by multiplying ${P}'$ with specific mapping matrix $M$ for each facet separately. $M$ is mentioned in Sec. 2.2 as a $9 \times k$ dimensional matrix. The mapping for a facet can be represented as:
\begin{equation}
  {Q}' = M {P}' = 
\begin{bmatrix}
m_{1, 1} &m_{1, 2} &\cdots &m_{1, k} \\
m_{2, 1} &m_{2, 2} &\cdots &m_{2, k} \\
\vdots &\vdots &\vdots &\vdots \\
m_{9, 1} &m_{9, 2} &\cdots &m_{9, k} \\
\end{bmatrix}
\begin{bmatrix}
p_1  \\ 
p_2 \\ 
\vdots  \\
p_k
\end{bmatrix}
=
\begin{bmatrix}
q_1  \\ 
q_2 \\ 
\vdots  \\
q_9
\end{bmatrix}.
\end{equation}
Here the transformation matrix can be obtained by reshaping ${Q}'$ to a $3 \times 3$ dimensional matrix.

Finally, vertex positions can be solved from the list of deformations by a more efficient method proposed by Sumner et al. \cite{sumner_deformation_2004}, which reformulates the deformation transfer optimization problem in terms of vertex positions.

\section{Evaluation}

\begin{figure}
\includegraphics[width=0.9\columnwidth]{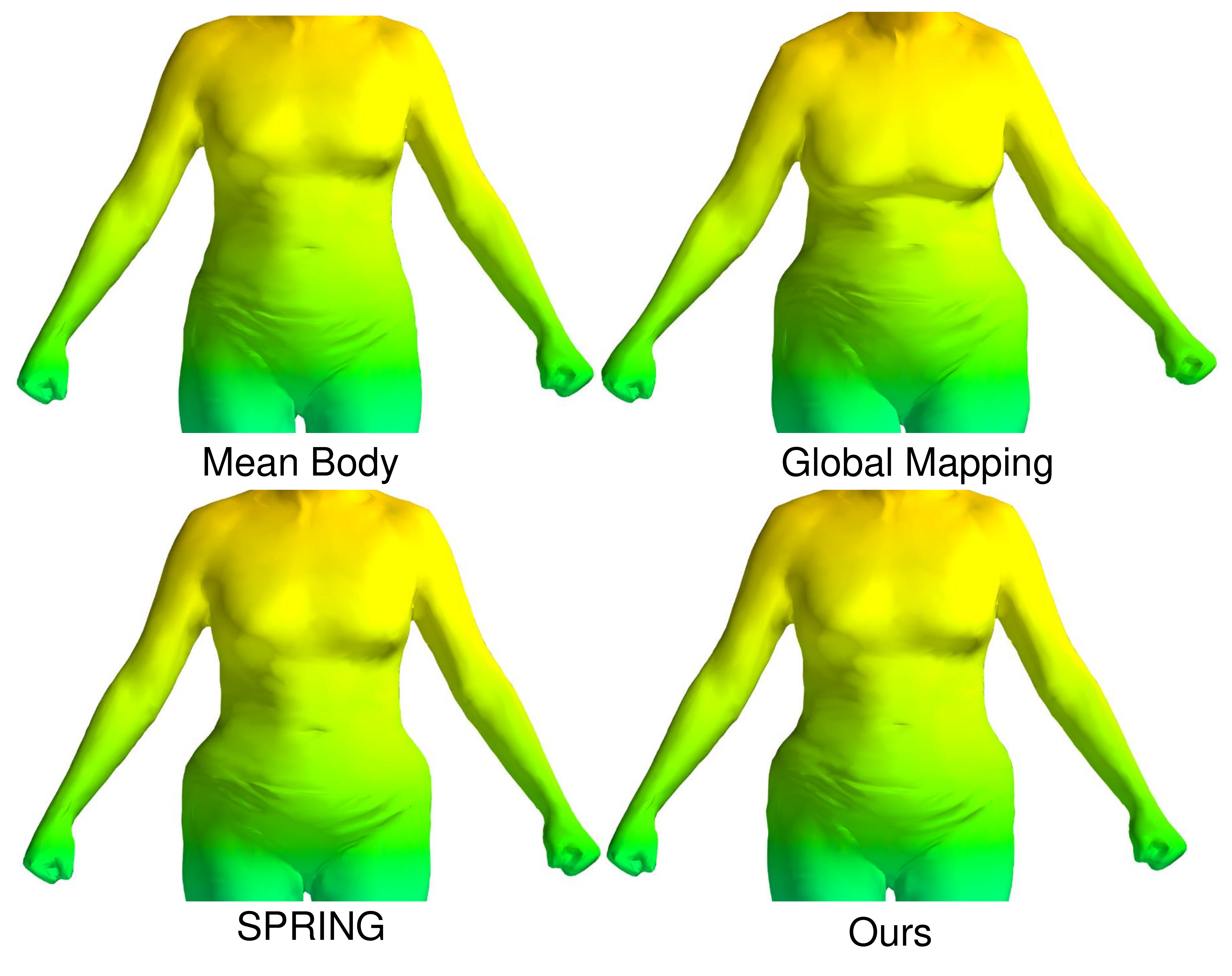}
\caption{Results generated by three models when the number of gluteal hip is increased by $12cm$.} 
\label{Fig:gluteal}
\end{figure}

Here we conduct extensive objective and subjective evaluations by anthropomorphic evaluation and a user study for our system.

We also trained state-of-art global mapping method and the SPRING model for comparison. The global mapping method learns the linear regression on anthropometric parameters and PCA coefficients of deformation matrices \cite{allen_space_2003}. As for the SPRING model, we conduct a survey on the relationship between each rigid part and anthropometric parameters, then we bind the relevant parameters to each part according to the analysis of the results of the survey.

These models are trained on the pose-independent dataset published by Y. Yang et al. \cite{yang_semantic_2014}. The dataset contains $1531$ female meshes and $1517$ male meshes. The resolution of each mesh is $12500$ vertices and $25000$ facets.


\subsection{Anthropomorphic evaluation}

Since anthropometric measurements are important in applications such as clothing sizing and our system generates 3D body meshes from anthropometric parameters, the error evaluation in such anthropometric parameters is necessary. We conduct an anthropomorphic error evaluation by calculating the mean absolute errors (MAE) in anthropometric measurements of reconstructed body mesh using a similar method used by Streuber et al. \cite{streuber_body_2016}. 

Taking the original anthropometric parameters of the bodies in the dataset as input, our experiments reconstruct the bodies with the global mapping method, the SPRING model and our system. To evaluate the accuracy of these three methods, we extract the anthropometric measurements from the generated meshes by calculating the point-to-point distance with sets of control points. Then for each model we calculate the mean absolute error between the measurements of generated meshes and those from the dataset. 

\autoref{table:mae} shows the mean absolute errors of reconstructing body meshes using 19 parameters with three models. Evaluation results shows that our proposed model outperforms the state-of-the-art approaches in terms of mean reconstruction error.


\begin{figure}
\includegraphics[width=0.9\columnwidth]{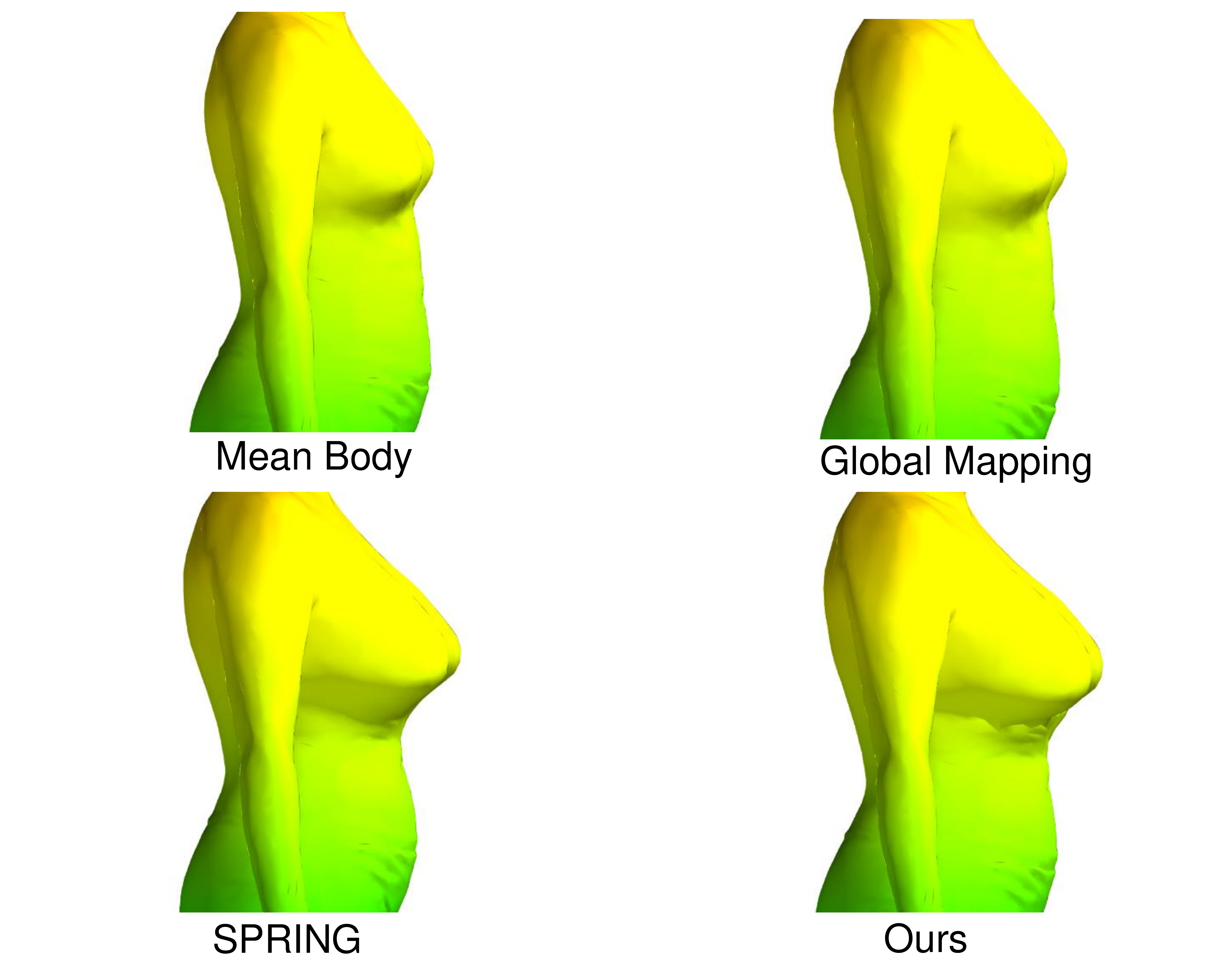}
\caption{Results generated by three models when the number of chest circumstance is increased by $10cm$.}
\label{Fig:chest}
\end{figure}

\begin{table}
\caption{Reconstruction errors of anthropomorphic evaluation. Mean absolute errors (MAE) for anthropometric parameters used in models: length error in mm, weight error in kg, and the numbers are listed as female/male. (The smaller, the better)} 
\label{table:mae}
\begin{tabular}{llll} 
    \toprule
    Parameter &global \cite{allen_space_2003} &SPRING \cite{yang_semantic_2014} & ours \\ 
    \midrule
    Weight                       & 1.4/1.4    & 1.2/1.4     & \underline{\textbf{0.9/0.9}} \\ 
    Height                       & 15.1/16.6  & 12.2/13.4   & \underline{\textbf{4.5/4.7}} \\ 
    neck                         & 8.5/11     & 2.9/3.5     & 3.2/4.0 \\ 
    chest                        & 28.1/25.6  & 11/11.8     & \underline{\textbf{10.3/11.4}}  \\ 
    belly button waist           & 23.8/25.0  & 10.3/13.3   & 10.7/14.1 \\ 
    gluteal hip                  & 20.6/22.3  & 9.01/9.9    & 9.5/10.4 \\ 
    neck shoulder elbow wrist    & 11.4/11.7  & 7.9/7.5     & \underline{\textbf{7.3/6.8}} \\ 
    crotch knee floor            & 15.4/15.4  & 10.4/9.2    & \underline{\textbf{10.4/9.1}}  \\ 
    across back shoulder neck    & 11.5/11.8  & 3.3/3.6     & 4.7/4.7 \\ 
    neck to gluteal hip          & 15.1/18.7  & 10.5/12.6   & \underline{\textbf{9.7/12}}  \\ 
    natural waist                & 23/23.2    & 10.5/11.9   & 10.6/12.3  \\ 
    max. hip                     & 21.7/21    & 10.1/9.3    & 10.3/9.7  \\ 
    natural waist rise           & 28.6/33.67 & 16.4/21.7   & 16.7/22.4  \\ 
    shoulder to midhand          & 8.3/9.0    & 3.8/4.1     & 3.8/4.3 \\ 
    upper arm                    & 10.5/11.3  & 5.4/6.8     & \underline{\textbf{4.8/5.9}} \\ 
    wrist                        & 4.5/4.9    & 2.4/2.1     & \underline{\textbf{2.4/2.1}} \\ 
    outer natural waist to floor & 14.4/12.9  & 9.9/10      & \underline{\textbf{9.1/8.4}} \\ 
    knee                         & 9.9/8.4    & 4.5/3.5     & 4.7/3.6 \\ 
    max. thigh                   & 18.5/18.2  & 16.1/14     & \underline{\textbf{14.7/12.7}}  \\
    length average error         & 16.1/16.7  & 8.7/9.3     & \underline{\textbf{8.3/8.8}} \\
    \bottomrule
\hline\end{tabular}
\end{table} 


\subsection{User study}

\begin{figure}
\center
\includegraphics[width=\columnwidth]{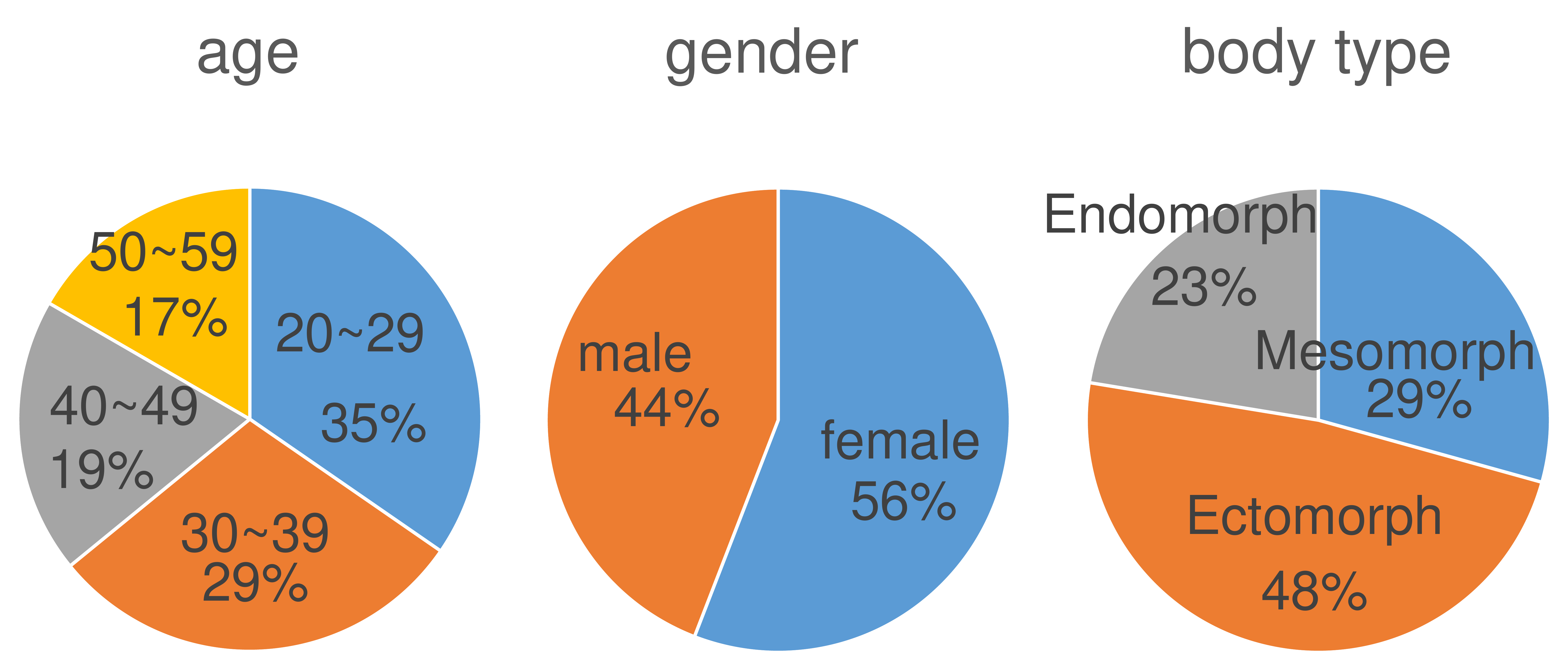}
\caption{The age distribution, gender distribution and body type distribution of volunteers. [Best viewed in color]} 
\label{Fig:user_statistic}
\end{figure}

Anthropomorphic evaluation does not always reflect the performance of shaping models. Sometimes the body meshes constructed with lower error do not look like the subject of interest, and vice versa. Here we adjust the value of a certain parameter from mean shape to see whether the change in the mesh is similar as we expected, and we also conduct a user study on these three models from 68 volunteers. The age distribution, gender distribution and body type (i.e. Mesomorph, Ectomorph, and Endomorph \cite{sheldon1940varieties}) distribution are shown in \autoref{Fig:user_statistic}.

We increase the number of gluteal hip circumstance by $12cm$. We analyze the results generated by different models in \autoref {Fig:gluteal}, and we find that the upper part of the result generated by the global mapping method changes a lot including chest size, belly button waist size etc. Our model produces a similar result with the one generated by SPRING model. Specifically, when the value of gluteal hip circumstance is increased, our model can keep the irrelevant parts (e.g., chest part, thigh part, shoulder part etc.) unchanged.

When the value of chest circumstance is increased by $10cm$, we observe from \autoref{Fig:chest} that the global mapping method is unable to reshape the 3D body mesh as expected. The result generated by the global mapping method seems unchanged compared with the mean body shape. The results generated by our system and the SPRING model both change significantly in the chest part of body, which outperforms the global mapping method dramatically.

\begin{figure}
\center
\includegraphics[width=\columnwidth]{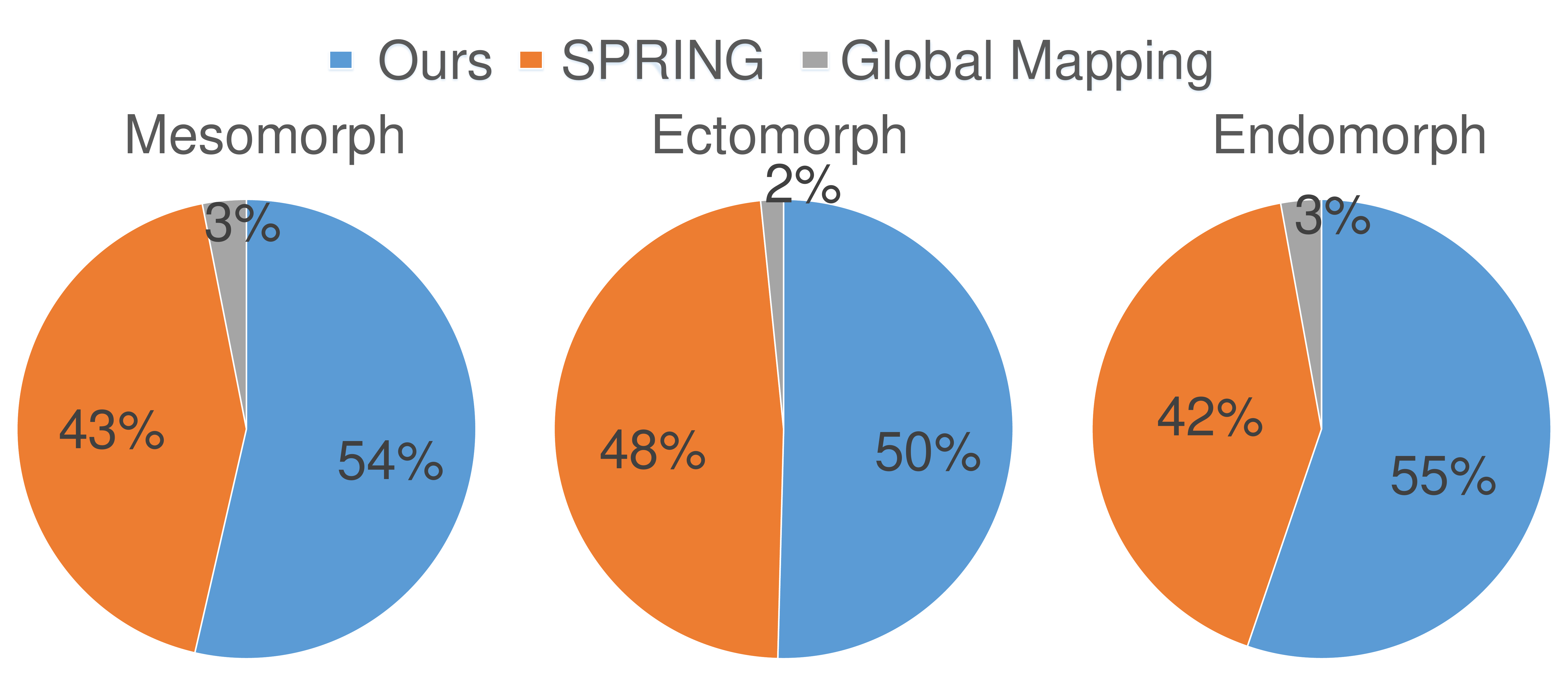}
\caption{The satisfaction survey among three body types. [Best viewed in color]} 
\label{Fig:user_study}
\end{figure}

We also conduct a user study on these three models. 68 volunteers are invited to experience these models. Volunteers are asked which model can generates the most resembling body mesh after experiencing all three models. The results shown in \autoref{Fig:user_study} according to three body types proves the advantages of our system.

\section{Conclusion}
In this paper, we have design a user-friendly and accurate system for 3D human body reshaping with limited anthropometric parameters by leveraging MICE technique for imputation and proposing a feature-selection-based local mapping method for shape modeling. The feature-selection-based local mapping method we proposesd here can select the most relevant parameters for each facet automatically for linear regression learning, which eliminates heavy human efforts for utilizing topology information of body shape, and thus a more approximate body mesh can be obtained. We also conduct anthropomorphic evaluation and a user study for our system. The results compared with other methods shows the advantages of our system. In future work, we plan to combine local methods with global methods for a faster and more accurate system. 


\begin{acks}
This work is partially supported by NSF of China under Grant 61672548, U1611461, and the Guangzhou Science and Technology Program, China, under Grant 201510010165.
\end{acks}

\bibliographystyle{ACM-Reference-Format}
\bibliography{sigproc.bib} 

\end{document}